\documentclass[twoside,11pt]{article}

\usepackage{jmlr2e}
\usepackage{booktabs}
\usepackage{float}
\usepackage{wrapfig}
\usepackage{graphicx}
\usepackage{multirow}
\usepackage{caption}
\usepackage{pifont}
\newcommand{\cmark}{\ding{51}}%
\usepackage[table]{xcolor}
\usepackage{color}
\usepackage{lastpage}

\newcommand*\samethanks[1][\value{footnote}]{\footnotemark[#1]}

\newcommand{\CC}[1]{\cellcolor{#1}}
\definecolor{gray}{rgb}{0.83, 0.83, 0.83}

\usepackage{xcolor}

\jmlrheading{23}{2022}{1-\pageref{LastPage}}{9/21; Revised 12/21}{1/22}{21-1155}{Victor G. Turrisi da Costa, Enrico Fini, Moin Nabi, Nicu Sebe, and Elisa Ricci}

\ShortHeadings{\texttt{solo-learn}: A Library of Self-supervised Methods for Visual Representation Learning}{Turrisi da Costa, Fini, Nabi, Sebe, and Ricci}
\firstpageno{1}
\begin{document}

\title{\texttt{solo-learn}: A Library of Self-supervised Methods\\for Visual Representation Learning}

\author{\name Victor G. Turrisi da Costa\thanks{Victor G. Turrisi da Costa and Enrico Fini contributed equally.} \email vg.turrisidacosta@unitn.it \\
       \addr University of Trento - Trento, Italy
       \AND
       \name Enrico Fini\samethanks \email enrico.fini@unitn.it \\
       \addr University of Trento - Trento, Italy
       \AND
       \name Moin Nabi \email m.nabi@sap.com \\ \addr SAP AI Research - Berlin, Germany       
       \AND
       \name Nicu Sebe \email niculae.sebe@unitn.it \\
       \addr University of Trento - Trento, Italy
       \AND
       \name Elisa Ricci \email e.ricci@unitn.it \\
       \addr University of Trento and Fondazione Bruno Kessler - Trento, Italy
}

\editor{Alexandre Gramfort}
\maketitle

\begin{abstract}

This paper presents \texttt{solo-learn}, a library of self-supervised methods for visual representation learning. Implemented in Python, using Pytorch and Pytorch lightning, the library fits both research and industry needs by featuring distributed training pipelines with mixed-precision, faster data loading via Nvidia DALI, online linear evaluation for better prototyping, and many additional training tricks.  Our goal is to provide an easy-to-use library comprising a large amount of Self-supervised Learning (SSL) methods, that can be easily extended and fine-tuned by the community. \texttt{solo-learn} opens up avenues for exploiting large-budget SSL solutions on inexpensive smaller infrastructures and seeks to democratize SSL by making it accessible to all.
The source code is available at \url{https://github.com/vturrisi/solo-learn}.

\end{abstract}


\section{Introduction}
Deep networks trained with large annotated datasets have shown stunning capabilities in the context of computer vision. However, the need for human supervision is a strong limiting factor. Unsupervised learning aims to mitigate this issue by training models from unlabeled datasets. The most prominent paradigm for unsupervised visual representation learning is Self-supervised Learning (SSL), where the intrinsic structure of the data provides supervision for the model. Recently, the scientific community devised increasingly effective SSL methods that match or surpass the performance of supervised methods. Nonetheless, implementing and reproducing such works turns out to be complicated. Official repositories of state-of-the-art SSL methods have very heterogeneous implementations or no implementation at all. Although a few SSL libraries \citep{goyal2021vissl, susmelj2020lightly} are available, they assume that larger-scale infrastructures are available or they lack some recent methods. When approaching SSL, it is hard to find a platform for experiments that allows running all current state of the art methods with low engineering effort and at the same time is effective and straightforward to train. This is especially problematic because, while the SSL methods seem simple on paper, replication of published results can involve a huge time and effort from researchers. Sometimes official implementations of SSL methods are available, however, releasing standalone packages (often incompatible with each other) is not sufficient for the fast-paced progress in research and emerging real-world applications.
There is no toolbox offering a genuine off-the-shelf catalog of state-of-the-art SSL techniques that is computationally efficient, which is essential for in-the-wild experimentation.

To address these problems, we present \texttt{solo-learn}, an open-source framework that provides standardized implementations for a large number of state-of-the-art SSL methods.
We believe \texttt{solo-learn} will enable a trustworthy and reproducible comparison between the state of the art methods.
The code that powers the library is written in Python, using Pytorch~\citep{pytorch} and Pytorch Lightning(PL)~\citep{falcon2019pytorch} as back-ends and Nvidia DALI\footnote{https://github.com/NVIDIA/DALI} for fast data loading, and supports more modern methods than related libraries.
The library is highly modular and can be used as a complete pipeline, from training to evaluation, or as standalone modules.

\section{The \texttt{solo-learn} Library: An Overview}

Currently, we are witnessing an explosion of works on SSL methods for computer vision.
Their underlying idea is to unsupervisedly learn feature representations by enforcing similar feature representations across multiple views from the same image while enforcing diverse representations for other images.
To help researchers have a common testbed for reproducing different results, we present \texttt{solo-learn}, which is a library of self-supervised methods for visual representation learning.
The library is implemented in Pytorch, providing state-of-the-art self-supervised methods, distributed training pipelines with mixed-precision, faster data loading, online linear evaluation for better prototyping, and many other training strategies and tricks presented in recent papers.
{\color{black}We also provide an easy way to use the pre-trained models for object detection, via DetectronV2 \citep{wu2019detectron2}.}
Our goal is to provide an easy-to-use library that can be easily extended by the community, while also including additional features that make it easier for researchers and practitioners to train on smaller infrastructures.

\subsection{Self-supervised Learning Methods}

We implemented 13 state-of-the-art methods, namely, Barlow Twins \citep{zbontar2021barlow}, BYOL \citep{grill2020bootstrap}, DeepCluster V2 \citep{caron2021unsupervised}, DINO \citep{caron2021emerging}, MoCo V2+ \citep{chen2020improved}, NNCLR \citep{dwibedi2021little}, ReSSL \citep{zheng2021ressl}, SimCLR \citep{chen2020simple}, Supervised Contrastive Learning \citep{khosla2020supervised}, SimSiam \citep{chen2020exploring}, SwAV \citep{caron2021unsupervised}, VICReg \citep{bardes2021vicreg} and W-MSE \citep{ermolov2021whitening}.

\subsection{Architecture}

\begin{figure*}[t]
    \centering
    \includegraphics[width=0.95\textwidth]{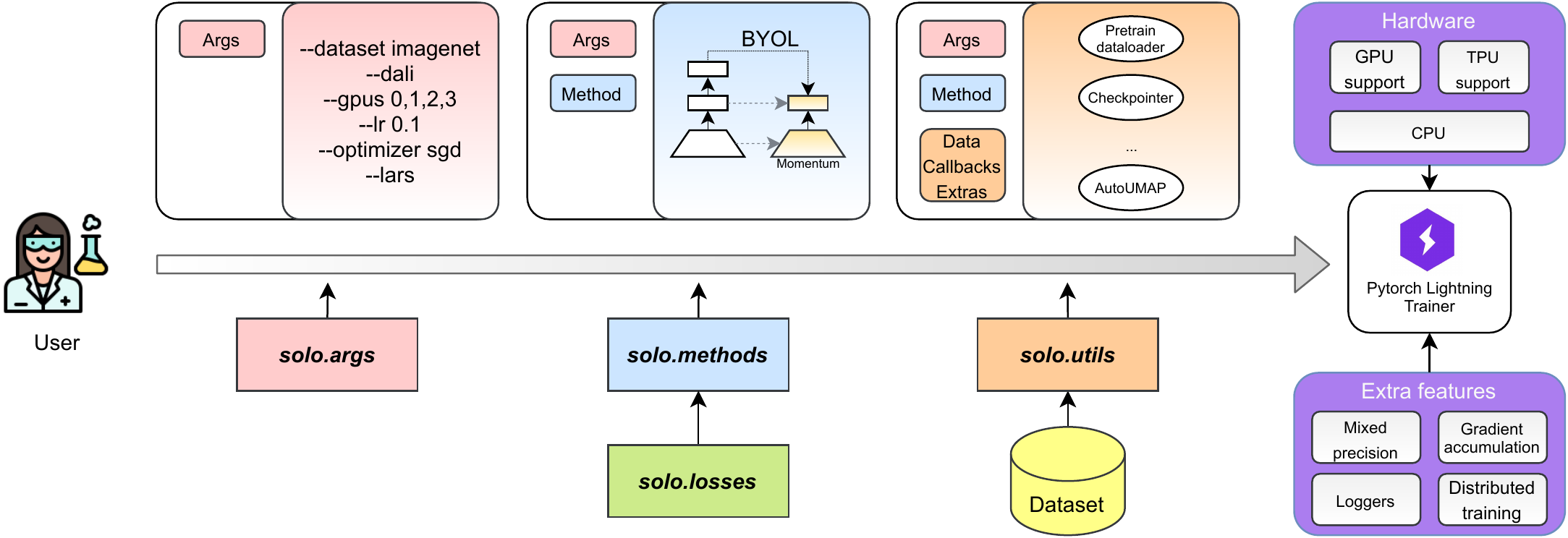}
    \caption{Overview of \texttt{solo-learn}.}
    \label{fig:overview}
    \vspace{-5mm}
\end{figure*}

In Figure \ref{fig:overview}, we present an overview of how a training pipeline with \texttt{solo-learn} is carried out.
In the bottom, we show the packages and external data at each step, while at the top, we show all the defined variables on the left and an example of the newest defined variable on the right.
First, the user interacts with {\color{red}\texttt{solo.args}}, a subpackage that is responsible for handling all the parameters selected by the user and providing automatic setup.
Then, {\color{blue}\texttt{solo.methods}} interacts with {\color{green}\texttt{solo.losses}} to produce the selected self-supervised method.
While {\color{blue}\texttt{solo.methods}} contains all implemented methods, {\color{green}\texttt{solo.losses}} contains the loss functions for each method.
Afterwards, {\color{orange}\texttt{solo.utils}} handles external data to produce the pretrain dataloader, which contains all the transformation pipelines, model checkpointer, automatic UMAP visualization of the features, {\color{black}other backbone networks, such as ViT \citep{dosovitskiy2020vit} and Swin \citep{liu2021Swin}}, and many other utility functionalities.
Lastly, this is given to a PL trainer, which provides hardware support and extra functionality, such as, distributed training, automatic logging results, mixed precision and much more.
We note that although we show all subpackages working together, they can be used in a standalone fashion with minor modifications.
{\color{black}Apart from that, we have documentations in the folder {\color{purple}\textbf{docs}}, downstream tasks in {\color{purple}\textbf{downstream}}, unit tests in {\color{purple}\textbf{tests}} and pretrained models in {\color{purple}\textbf{zoo}}.}

\subsection{Comparison to Related Libraries}

The most related libraries to ours are VISSL \citep{goyal2021vissl} and Lightly \citep{susmelj2020lightly}, which lack some of our key features.
First, we support more modern SSL methods, such as BYOL, NNCLR, SimSiam, VICReg, W-MSE and others.
Second, we target researchers with fewer resources, namely from 1 to 8 GPUs, allowing much faster data loading via DALI.
Lastly, we provide additional utilities, such as automatic linear evaluation, support to custom datasets and automatically generating UMAP~\citep{2018arXivUMAP} visualizations of the features during training.

\section{Experiments}
\paragraph{Benchmarks.} We benchmarked the available SSL methods on CIFAR-10~\citep{cifar}, CIFAR-100~\cite{cifar} and ImageNet-100~\citep{deng2009imagenet} and made public the pretrained checkpoints.
For Barlow Twins, BYOL, MoCo V2+, NNCLR, SimCLR and VICReg, hyperparameters were heavily tuned, reaching higher performance than reported on original papers or third-party results.
Tab.~\ref{tab:results} presents the top-1 and top-5 accuracy values for the online linear evaluation. For ImageNet-100, traditional offline linear evaluation is also reported. We also compare with the results reported by Lightly in Tab.~\ref{tab:lightly}.

\paragraph{Nvidia DALI vs traditional data loading.} We compared the training speeds and memory usage of using traditional data loading via Pytorch Vision\footnote{https://github.com/pytorch/vision} against data loading with DALI.
For consistency, we ran three different methods (Barlow Twins, BYOL and NNCLR) for 20 epochs on ImageNet-100. Tab.~\ref{tab:dali} presents these results.

\begin{table}[!ht]
        \centering
        \small
        \captionsetup{width=0.999\textwidth}
        \caption{Online linear evaluation accuracy on CIFAR-10, CIFAR-100 and ImageNet-100. In brackets, offline linear evaluation accuracy is also reported for ImageNet-100.}
            \begin{tabular}{lcccccccc}
                \toprule
                \multirow{2}[2]{*}{\textbf{Method}} & \multicolumn{2}{c}{\textbf{CIFAR-10}} & \multicolumn{2}{c}{\textbf{CIFAR-100}} & \multicolumn{2}{c}{\textbf{ImageNet-100}} \\
                \cmidrule(lr){2-3} \cmidrule(lr){4-5} \cmidrule(lr){6-7}
                & Acc@1 & Acc@5 & Acc@1 & Acc@5 & Acc@1 & Acc@5 \\
                \midrule
                Barlow Twins & 92.10 & 99.73 & 70.90 & 91.91 & 80.38 (80.16) & 95.28 (95.14) \\
                BYOL & 92.58 & 99.79 & 70.46 & 91.96 & 80.16 (80.32) & 94.80 (94.94)\\ 
                DeepCluster V2 & 88.85 & 99.58 & 63.61 & 88.09 & 75.36 (75.40) & 93.22 (93.10) \\ 
                DINO & 89.52 & 99.71 & 66.76 & 90.34 & 74.84 (74.92) & 92.92 (92.78) \\ 
                MoCo V2+ & 92.94 & 99.79 & 69.89 &  91.65 & 78.20 (79.28) & 95.50 (95.18) \\ 
                NNCLR & 91.88 & 99.78 & 69.62 & 91.52 & 79.80 (80.16) & 95.28 (95.28) \\ 
                ReSSL & 90.63 & 99.62 & 65.92 & 89.73 & 76.92 (78.48) & 94.20 (94.24) \\ 
                SimCLR & 90.74 & 99.75 & 65.78 & 89.04 & 77.04 (77.48) & 94.02 (93.42) \\ 
                Simsiam & 90.51 & 99.72 & 66.04 & 89.62 & 74.54 (78.72) & 93.16 (94.78) \\ 
                SwAV & 89.17 & 99.68 & 64.88 & 88.78 & 74.04 (74.28) & 92.70 (92.84) \\ 
                VICReg & 92.07 & 99.74 & 68.54 & 90.83 & 79.22 (79.40) & 95.06 (95.02)\\
                W-MSE & 88.67 & 99.68 & 61.33 & 87.26 & 67.60 (69.06) & 90.94 (91.22) \\
                \bottomrule
            \end{tabular}
        \label{tab:results}
        \vspace{-12pt}
\end{table}

\begin{table}[!h]
    \scriptsize
    \centering
    \begin{minipage}[b]{0.68\textwidth}\centering
    \captionsetup{width=0.95\textwidth}
    \caption{Speed and memory comparison with and without DALI on ImageNet-100.}
    \begin{tabular}{lccccc}
        \toprule
        \textbf{Method} & \textbf{
DALI} & \textbf{20 epochs} & \textbf{1 epoch} & \textbf{Speedup} & \textbf{Memory} \\ 
        \midrule
        \multirow{2}{*}{\parbox{1.1pt}{Barlow\\Twins}} & & 1h 38m 27s & 4m 55s & - & 5097 MB  \\
        & \CC{gray}\cmark & \CC{gray}43m 2s & \CC{gray}2m 10s & \CC{gray}56\% & \CC{gray}9292 MB  \\
        \midrule
        \multirow{2}{*}{BYOL} & & 1h 38m 46s & 4m 56s & - &5409 MB  \\
        & \CC{gray}\cmark & \CC{gray}50m 33s & \CC{gray}2m 31s & \CC{gray}49\% & \CC{gray}9521 MB  \\
        \midrule
        \multirow{2}{*}{NNCLR} & & 1h 38m 30s & 4m 55s & - & 5060 MB  \\
        & \CC{gray}\cmark & \CC{gray}42m 3s & \CC{gray}2m 6s & \CC{gray}64\% & \CC{gray}9244 MB  \\
        \bottomrule
    \end{tabular}
    \label{tab:dali}
    \end{minipage}
    \begin{minipage}[b]{0.30\textwidth}\centering
    \captionsetup{width=0.95\textwidth}
    \caption{Comparison with Lightly on CIFAR10.}
    \label{tab:lightly}
    \begin{tabular}{lcc}
        \toprule
        \textbf{Method} & \textbf{Ours} & \textbf{Lightly} \\ 
        \midrule
        SimCLR & \CC{gray}90.74 & 89.0 \\
        MoCoV2+ & \CC{gray}92.94 & 90.0   \\
        SimSiam& \CC{gray}90.51 & 91.0   \\
        \bottomrule
    \end{tabular}
    \end{minipage}
    \vspace{-10pt}
\end{table}

\section{Conclusion}
Here, we presented \texttt{solo-learn}, a library of self-supervised methods for visual representation learning, providing state-of-the-art self-supervised methods in Pytorch.
The library supports distributed training, fast data loading and provides many utilities for the end-user, such as online linear evaluation for better prototyping and faster development, many training tricks, and visualization techniques.
We are continuously adding new SSL methods, improving usability, documents, and tutorials.
Finally, we welcome contributors to help us at  \url{https://github.com/vturrisi/solo-learn}.

\noindent \textbf{Acknowledgments.} This work was supported by a joint project under Grant No. JQ18012, by the EU H2020 AI4Media No. 951911 project and the European Institute of Innovation \& Technology (EIT).


\vskip 0.2in
\bibliography{sample}

\begin{thebibliography}{22}
\providecommand{\natexlab}[1]{#1}
\providecommand{\url}[1]{\texttt{#1}}
\expandafter\ifx\csname urlstyle\endcsname\relax
  \providecommand{\doi}[1]{doi: #1}\else
  \providecommand{\doi}{doi: \begingroup \urlstyle{rm}\Url}\fi

\bibitem[Bardes et~al.(2021)Bardes, Ponce, and LeCun]{bardes2021vicreg}
Adrien Bardes, Jean Ponce, and Yann LeCun.
\newblock Vicreg: Variance-invariance-covariance regularization for
  self-supervised learning.
\newblock \emph{arXiv:2105.04906}, 2021.

\bibitem[Caron et~al.(2020)Caron, Misra, Mairal, Goyal, Bojanowski, and
  Joulin]{caron2021unsupervised}
Mathilde Caron, Ishan Misra, Julien Mairal, Priya Goyal, Piotr Bojanowski, and
  Armand Joulin.
\newblock Unsupervised learning of visual features by contrasting cluster
  assignments.
\newblock In \emph{NeurIPS}, 2020.

\bibitem[Caron et~al.(2021)Caron, Touvron, Misra, Jégou, Mairal, Bojanowski,
  and Joulin]{caron2021emerging}
Mathilde Caron, Hugo Touvron, Ishan Misra, Hervé Jégou, Julien Mairal, Piotr
  Bojanowski, and Armand Joulin.
\newblock Emerging properties in self-supervised vision transformers.
\newblock \emph{arXiv:2104.14294}, 2021.

\bibitem[Chen et~al.(2020{\natexlab{a}})Chen, Kornblith, Norouzi, and
  Hinton]{chen2020simple}
Ting Chen, Simon Kornblith, Mohammad Norouzi, and Geoffrey Hinton.
\newblock A simple framework for contrastive learning of visual
  representations.
\newblock In \emph{ICML}, 2020{\natexlab{a}}.

\bibitem[Chen and He(2021)]{chen2020exploring}
Xinlei Chen and Kaiming He.
\newblock Exploring simple siamese representation learning.
\newblock In \emph{CVPR}, 2021.

\bibitem[Chen et~al.(2020{\natexlab{b}})Chen, Fan, Girshick, and
  He]{chen2020improved}
Xinlei Chen, Haoqi Fan, Ross Girshick, and Kaiming He.
\newblock Improved baselines with momentum contrastive learning.
\newblock \emph{arXiv:2003.04297}, 2020{\natexlab{b}}.

\bibitem[Deng et~al.(2009)Deng, Dong, Socher, Li, Li, and
  Fei-Fei]{deng2009imagenet}
Jia Deng, Wei Dong, Richard Socher, Li-Jia Li, Kai Li, and Li~Fei-Fei.
\newblock Imagenet: A large-scale hierarchical image database.
\newblock In \emph{2009 IEEE conference on computer vision and pattern
  recognition}, 2009.

\bibitem[Dosovitskiy et~al.(2021)Dosovitskiy, Beyer, Kolesnikov, Weissenborn,
  Zhai, Unterthiner, Dehghani, Minderer, Heigold, Gelly, Uszkoreit, and
  Houlsby]{dosovitskiy2020vit}
Alexey Dosovitskiy, Lucas Beyer, Alexander Kolesnikov, Dirk Weissenborn,
  Xiaohua Zhai, Thomas Unterthiner, Mostafa Dehghani, Matthias Minderer, Georg
  Heigold, Sylvain Gelly, Jakob Uszkoreit, and Neil Houlsby.
\newblock An image is worth 16x16 words: Transformers for image recognition at
  scale.
\newblock \emph{ICLR}, 2021.

\bibitem[Dwibedi et~al.(2021)Dwibedi, Aytar, Tompson, Sermanet, and
  Zisserman]{dwibedi2021little}
Debidatta Dwibedi, Yusuf Aytar, Jonathan Tompson, Pierre Sermanet, and Andrew
  Zisserman.
\newblock With a little help from my friends: Nearest-neighbor contrastive
  learning of visual representations.
\newblock \emph{arXiv:2104.14548}, 2021.

\bibitem[Ermolov et~al.(2021)Ermolov, Siarohin, Sangineto, and
  Sebe]{ermolov2021whitening}
Aleksandr Ermolov, Aliaksandr Siarohin, Enver Sangineto, and Nicu Sebe.
\newblock Whitening for self-supervised representation learning.
\newblock In \emph{ICML}, 2021.

\bibitem[Goyal et~al.(2021)Goyal, Duval, Reizenstein, Leavitt, Xu, Lefaudeux,
  Singh, Reis, Caron, Bojanowski, Joulin, and Misra]{goyal2021vissl}
Priya Goyal, Quentin Duval, Jeremy Reizenstein, Matthew Leavitt, Min Xu,
  Benjamin Lefaudeux, Mannat Singh, Vinicius Reis, Mathilde Caron, Piotr
  Bojanowski, Armand Joulin, and Ishan Misra.
\newblock Vissl.
\newblock \emph{GitHub. Note: https://github.com/facebookresearch/vissl}, 2021.

\bibitem[Grill et~al.(2020)Grill, Strub, Altch\'{e}, Tallec, Richemond,
  Buchatskaya, Doersch, Avila~Pires, Guo, Gheshlaghi~Azar, Piot, kavukcuoglu,
  Munos, and Valko]{grill2020bootstrap}
Jean-Bastien Grill, Florian Strub, Florent Altch\'{e}, Corentin Tallec, Pierre
  Richemond, Elena Buchatskaya, Carl Doersch, Bernardo Avila~Pires, Zhaohan
  Guo, Mohammad Gheshlaghi~Azar, Bilal Piot, koray kavukcuoglu, Remi Munos, and
  Michal Valko.
\newblock Bootstrap your own latent - a new approach to self-supervised
  learning.
\newblock In \emph{NeurIPS}, 2020.

\bibitem[Khosla et~al.(2020)Khosla, Teterwak, Wang, Sarna, Tian, Isola,
  Maschinot, Liu, and Krishnan]{khosla2020supervised}
Prannay Khosla, Piotr Teterwak, Chen Wang, Aaron Sarna, Yonglong Tian, Phillip
  Isola, Aaron Maschinot, Ce~Liu, and Dilip Krishnan.
\newblock Supervised contrastive learning.
\newblock \emph{NeurIPS}, 2020.

\bibitem[Krizhevsky et~al.(2009)Krizhevsky, Nair, and Hinton]{cifar}
Alex Krizhevsky, Vinod Nair, and Geoffrey Hinton.
\newblock Learning multiple layers of features from tiny images.
\newblock 2009.

\bibitem[Liu et~al.(2021)Liu, Lin, Cao, Hu, Wei, Zhang, Lin, and
  Guo]{liu2021Swin}
Ze~Liu, Yutong Lin, Yue Cao, Han Hu, Yixuan Wei, Zheng Zhang, Stephen Lin, and
  Baining Guo.
\newblock Swin transformer: Hierarchical vision transformer using shifted
  windows.
\newblock \emph{International Conference on Computer Vision (ICCV)}, 2021.

\bibitem[McInnes et~al.(2020)McInnes, Healy, and Melville]{2018arXivUMAP}
Leland McInnes, John Healy, and James Melville.
\newblock Umap: Uniform manifold approximation and projection for dimension
  reduction.
\newblock \emph{arXiv:1802.03426}, 2020.

\bibitem[Paszke et~al.(2019)Paszke, Gross, Massa, Lerer, Bradbury, Chanan,
  Killeen, Lin, Gimelshein, Antiga, Desmaison, Kopf, Yang, DeVito, Raison,
  Tejani, Chilamkurthy, Steiner, Fang, Bai, and Chintala]{pytorch}
Adam Paszke, Sam Gross, Francisco Massa, Adam Lerer, James Bradbury, Gregory
  Chanan, Trevor Killeen, Zeming Lin, Natalia Gimelshein, Luca Antiga, Alban
  Desmaison, Andreas Kopf, Edward Yang, Zachary DeVito, Martin Raison, Alykhan
  Tejani, Sasank Chilamkurthy, Benoit Steiner, Lu~Fang, Junjie Bai, and Soumith
  Chintala.
\newblock Pytorch: An imperative style, high-performance deep learning library.
\newblock In \emph{NeurIPS}, 2019.

\bibitem[Susmelj et~al.(2020)Susmelj, Heller, Wirth, Prescott, and
  et~al.]{susmelj2020lightly}
Igor Susmelj, Matthias Heller, Philipp Wirth, Jeremy Prescott, and Malte~Ebner
  et~al.
\newblock Lightly.
\newblock \emph{GitHub. Note: https://github.com/lightly-ai/lightly}, 2020.

\bibitem[Team(2019)]{falcon2019pytorch}
Pytorch Lightning~Development Team.
\newblock Pytorch lightning.
\newblock \emph{GitHub. Note:
  https://github.com/PyTorchLightning/pytorch-lightning}, 3, 2019.

\bibitem[Wu et~al.(2019)Wu, Kirillov, Massa, Lo, and
  Girshick]{wu2019detectron2}
Yuxin Wu, Alexander Kirillov, Francisco Massa, Wan-Yen Lo, and Ross Girshick.
\newblock Detectron2.
\newblock \url{https://github.com/facebookresearch/detectron2}, 2019.

\bibitem[Zbontar et~al.(2021)Zbontar, Jing, Misra, Lecun, and
  Deny]{zbontar2021barlow}
Jure Zbontar, Li~Jing, Ishan Misra, Yann Lecun, and Stephane Deny.
\newblock Barlow twins: Self-supervised learning via redundancy reduction.
\newblock In \emph{ICML}, 2021.

\bibitem[Zheng et~al.(2021)Zheng, You, Wang, Qian, Zhang, Wang, and
  Xu]{zheng2021ressl}
Mingkai Zheng, Shan You, Fei Wang, Chen Qian, Changshui Zhang, Xiaogang Wang,
  and Chang Xu.
\newblock Ressl: Relational self-supervised learning with weak augmentation.
\newblock \emph{arXiv:2107.09282}, 2021.

\end{thebibliography}

\end{document}